\documentclass[letterpaper, 10 pt, journal, twoside]{ieeetran}
\IEEEoverridecommandlockouts                             
\usepackage{color,soul}

\usepackage{graphicx} 
\usepackage{mathtools} 
\usepackage{siunitx}
\usepackage{threeparttable}
\usepackage{cite}
\usepackage{graphicx}
\usepackage{setspace}
\usepackage{acronym}
\usepackage{amsfonts}			
\usepackage{times}		
\usepackage{amssymb}
\usepackage{amsmath}
\usepackage{gensymb}
\usepackage{tabularx} 
\usepackage{multirow}
\usepackage{algorithm}
\usepackage{algorithmic}
\usepackage{booktabs}
\usepackage[T1]{fontenc}

\usepackage{float}
\usepackage{mwe}
\newcolumntype{L}{>{\centering\arraybackslash}m{3cm}}
\usepackage[thinc]{esdiff}
\usepackage[table]{xcolor}
\usepackage{textcomp}
\usepackage{hyperref}
\hypersetup{
    colorlinks=true,
    citecolor=black,
    linkcolor=black,
    filecolor=black,      
    urlcolor=black,
    }
\usepackage{url}
\usepackage{array,ragged2e}
\newcolumntype{P}[1]{>{\RaggedRight\arraybackslash}p{#1}}

\usepackage[font=small,skip=2pt]{caption}
\usepackage{textcomp, gensymb}
\usepackage{dblfloatfix}

\newcommand{\p}[1]{\medskip \noindent \textbf{{#1}.}}

\usepackage{microtype}
\renewcommand{\baselinestretch}{0.92}
\usepackage{etoolbox}
\newbool{showcolor}
\setbool{showcolor}{false}
\newcommand{\changed}[1]{%
  \ifbool{showcolor}{\textcolor{blue}{#1}}{#1}%
}

\title{\textit{Gotta Grow Fast}: Design and Benchmarking of a \\Tip Mount for \changed{High-Speed} Vine Robots}
\author{Antonio Alvarez Valdivia$^{1}$, Robert Reeve$^{1}$, Ankush Dhawan$^{1,2}$, Ciera McFarland$^{3}$, \\Chad Council$^{1}$, Margaret McGuinness$^{3}$, and Nathaniel Hanson$^{1}$

\thanks{Manuscript received: November 18, 2025; Revised April 3, 2026; Accepted June 3, 2026.}
\thanks{This paper was recommended for publication by Editor Yong-Lae Park upon evaluation of the Associate Editor and Reviewers' comments.}

\thanks{$^{1}$Antonio Alvarez Valdivia, Robert Reeve, Ankush Dhawan, Chad Council, and Nathaniel Hanson are with the Lincoln Laboratory, Massachusetts Institute of Technology, Lexington, Massachusetts, USA}
\thanks{$^{2}$Ankush Dhawan is with Stanford University, Stanford, California, USA}
\thanks{$^{3}$Ciera McFarland and Margaret McGuinness are with the University of Notre Dame, Notre Dame, Indiana, USA}
\thanks{Correspondence: {\tt\small nhanson2@mit.edu}}
\thanks{Digital Object Identifier (DOI): see top of this page.}

\thanks{
DISTRIBUTION STATEMENT A. Approved for public release. Distribution is unlimited.
This material is based upon work supported by the Department of the Air Force under Air Force Contract No. FA8702-15-D-0001 or FA8702-25-D-B002. Any opinions, findings, conclusions or recommendations expressed in this material are those of the author(s) and do not necessarily reflect the views of the Department of the Air Force. © 2026 Massachusetts Institute of Technology. Delivered to the U.S. Government with Unlimited Rights, as defined in DFARS Part 252.227-7013 or 7014 (Feb 2014). Notwithstanding any copyright notice, U.S. Government rights in this work are defined by DFARS 252.227-7013 or DFARS 252.227-7014 as detailed above. Use of this work other than as specifically authorized by the U.S. Government may violate any copyrights that exist in this work.}}
\begin{document}

\maketitle

\begin{abstract}
Soft, growing vine robots extend through tip eversion, a mechanism that enables navigation through cluttered environments. However, integrating cameras and other sensors at the tip is uniquely challenging because the material forming the tip is constantly renewed as the robot grows. This continual material turnover, combined with friction between internal layers, added tip weight, and fabric constriction, complicates sensor and tool mounting. These limitations hinder the deployment of vine robots for inspection and search tasks, where rapid growth while carrying tip-mounted sensors is essential.
In this work, we present a triangular roller tip mount that reduces internal resistance during growth by rolling rather than sliding against the robot body. The design was refined through iterative failure analysis, enabling, for the first time, consistent eversion on a TPU-coated ripstop nylon vine robot.
To quantitatively evaluate mount performance, we introduce a custom testbed that isolates tip mounting effects by measuring tail tension during eversion. Comparative experiments across multiple mount variants, including prior designs, show that our triangular roller mount achieves the lowest tail tension and most repeatable growth performance.
These results establish both a validated tip mount design and a repeatable benchmarking framework for advancing sensor and tool integration in soft growing robots. CAD for the mount and testbed is available at: \url{https://sprout-mitll.github.io/tip_mounts/}.
\end{abstract}


\section{Introduction}


\begin{figure}[!t]
    \centering
    \includegraphics[width = \columnwidth]{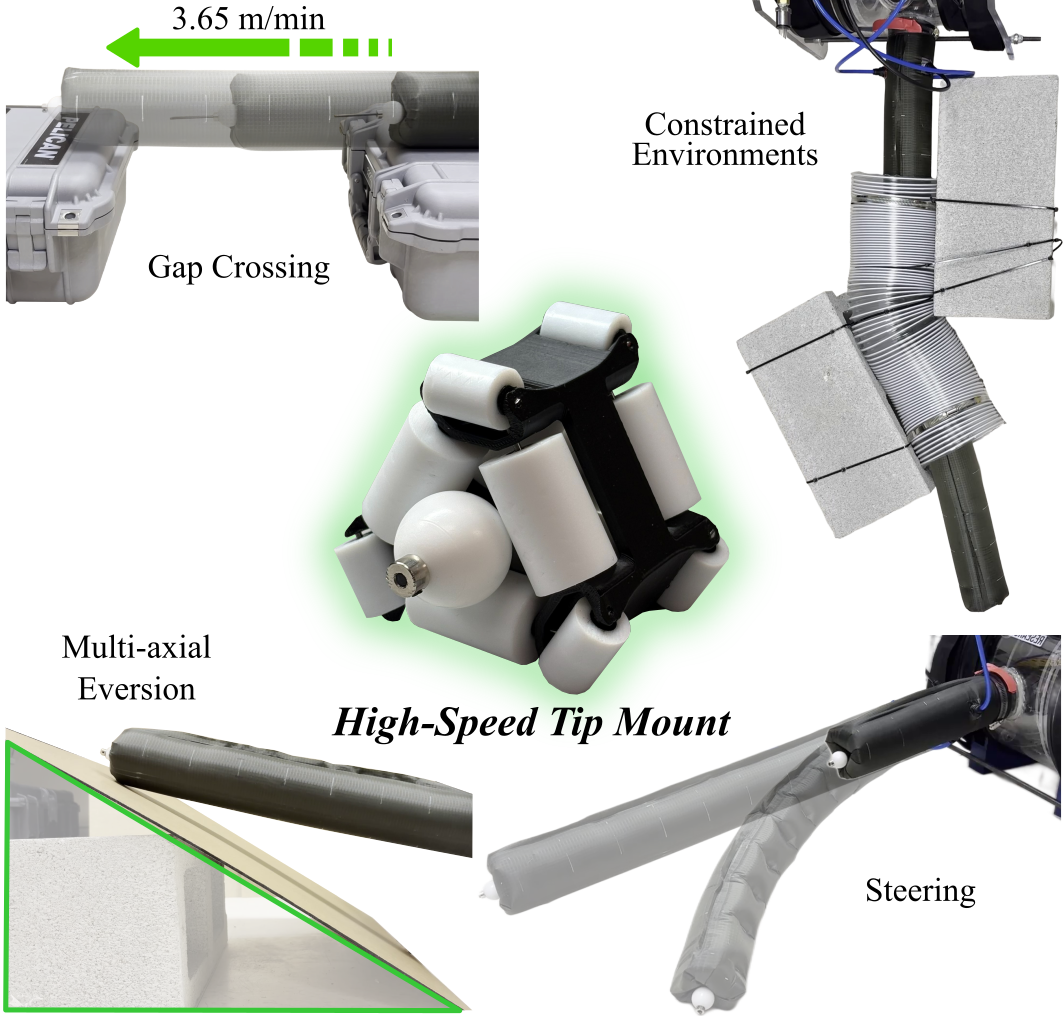}
    \caption{\changed{\textbf{High-speed tip mount for soft, growing robots.} Our triangular tip was validated on a 2.0 meter vine robot \cite{mcfarland2024field} with integrated pouch motors constructed from TPU-coated ripstop nylon. With 17.2 kPa in the main body, the robot is able to turn and grow at 3.65 m/min through confined spaces, over gaps, and up inclines (see figure for snapshots and supplementary video for full demonstrations).}}
    \label{fig:camera mount design}
    \vspace{-2.0em}
\end{figure}

\IEEEPARstart{R}{obots} play a crucial role in gathering information from hazardous environments. To be effective, they must not only traverse spaces but also carry the sensors needed for their tasks. For example, in search-and-rescue settings, cameras, microphones, and lights provide essential situational awareness for responders \cite{murphy2017disaster, murphy2004human}. In these operations, time is often the most critical variable; rapid deployment can mean the difference between locating a survivor and missing a rescue window. Soft robots must be capable of fast, reliable movement in order to become invaluable tools for first responders. Integrating necessary payloads while maintaining high speeds, however, remains a significant challenge, particularly for soft robots, whose structures are lightweight and compliant. 

Vine robots, soft, inflatable robots that extend by tip eversion \cite{blumenschein2020design,al2025tip}, have demonstrated their potential in exploration. They have been deployed in environments such as mock rubble piles \cite{der2021roboa, mcfarland2024field}, archaeological sites \cite{CoadRAM2020}, salamander burrows \cite{qin20253d}, and industrial piping systems \cite{heap2024large}. They can navigate through tight apertures \cite{HawkesScienceRobotics2017} and grow vertically for inspection \cite{CoadRAM2020}. These capabilities make them attractive for carrying sensors into confined or dangerous areas. To be viable in emergency response, such robots must not only reach hard-to-access regions but also do so quickly, maintaining growth speeds that match the urgency of their tasking.

Tip-mounted payloads introduce unique challenges: additional weight destabilizes the robot \cite{mcfarland2025onsteerability}, and rigid mounts prevent diameter constriction and induce friction at the tip, both of which can dramatically reduce growth speed and reliability. A variety of tip mounts have been proposed to address these issues \cite{jeong2020tip, heap2021soft, suulker2023soft, luong2019eversion, mishima2006development, luong2019eversion, CoadRAM2020, frias2023vine, heap2024large, der2021roboa}, yet each design involves tradeoffs. Many are heavy (increasing the risk of collapse under transverse loading) and come into contact with the environment during growth. Critically, few designs explicitly quantify how mount characteristics influence achievable growth rate, a key metric for time-sensitive missions. Prior work has not systematically benchmarked how tip mounts impact growth mechanics, leaving open the questions of how best to design mounts that balance weight, friction, and growth speed. 

In developing our own mount design, we noticed that many vine robots do not maintain a perfectly circular cross section during growth \cite{Przybylski2023_3D}. Instead, the body consistently unfolds (and folds back during retraction) along a triangular profile, giving the tip a three-sided geometry, which we attribute to the radial placement of series pouch motor actuators~\cite{CoadRAM2020} for directional steering. Leveraging this observation, we sought to reduce internal resistance to enable faster, more stable growth. This motivated our novel design of a tip mount with a triangular set of rollers (Fig.~\ref{fig:camera mount design}). By matching the mount geometry to the robot’s natural deformation, we sought to minimize internal resistance and improve consistency during growth, which is critical for sustaining high-speed extension over long distances. 

Building on this insight, we present the following contributions to the state of the art:
\begin{itemize}
    \item A lightweight, low-friction tip mount featuring a triangular roller assembly that minimizes internal resistance from higher-friction fabrics, enabling higher growth speeds for rapid deployment in time-critical scenarios.
    \item A benchmarking testbed that isolates the effect of tip mounts on growth mechanics by measuring tail tension during eversion, providing a direct measure of how mount design influences growth performance.
    \item A comparative study of multiple mount variants showing that our triangular roller mount achieves the lowest tail tension, fastest sustained growth rate, and the most consistent performance.
\end{itemize}

\section{Related Work}
Many tip mounts have been designed for vine robots in various environments. Here, we present an overview of existing designs and several qualities that are important for applications in urban search and rescue. All ten mounts and their qualities are summarized in Table~\ref{tab:tip_mounts}. We aim to design a mount that works on the higher-friction fabrics used for durable, field-ready vine robots, is low-profile with minimal environmental drag, is lightweight, is well-secured to the vine body, and allows for fast growth.

\begin{table}[b]
    \vspace{-1.0em}
    \centering
    \caption{\textbf{Characteristics of Existing Vine Robot Tip Mounts}}
    \label{tab:tip_mounts}
    \begingroup
    \scriptsize
    \setlength{\tabcolsep}{2pt}
    \renewcommand{\arraystretch}{1.0}

    \newcolumntype{Y}{>{\centering\arraybackslash}m{0.12\columnwidth}}
    \newcolumntype{Z}{>{\raggedright\arraybackslash}m{0.22\columnwidth}} 
    \newcolumntype{M}{>{\raggedright\arraybackslash}m{0.20\columnwidth}} 

    \begin{tabularx}{\columnwidth}{Z M Y Y Y Y}
        \toprule
        \textbf{Design} & \textbf{Body Material} & \textbf{Low Profile} & \textbf{Lightweight} & \textbf{Secure} & \textbf{Fast} \\
        \midrule
        Reeled Hard Cap~\cite{mishima2006development} & \cellcolor{green!20}Kevlar/TPU 
            & \cellcolor{red!20}No 
            & \cellcolor{red!20}No 
            & \cellcolor{green!20}Yes 
            & \cellcolor{yellow!20}-\\
        Magnets~\cite{luong2019eversion, stroopa2020icra, stroppa2023shared} & \cellcolor{red!20}Nylon/Silicone 
            & \cellcolor{green!20}Yes 
            & \cellcolor{green!20}Yes 
            & \cellcolor{red!20}No 
            & \cellcolor{yellow!20}-\\
        Hard Cap~\cite{CoadRAM2020} & \cellcolor{green!20}Nylon/TPU 
            & \cellcolor{red!20}No 
            & \cellcolor{red!20}No 
            & \cellcolor{red!20}No 
            & \cellcolor{green!20}Yes \\
        Tool~\cite{jeong2020tip} & \cellcolor{red!20}LDPE 
            & \cellcolor{red!20}No 
            & \cellcolor{red!20}No 
            & \cellcolor{green!20}Yes 
            & \cellcolor{green!20}Yes \\
        Internal~\cite{heap2021soft} & \cellcolor{red!20}Nylon/Silicone 
            & \cellcolor{green!20}Yes 
            & \cellcolor{green!20}Yes
            & \cellcolor{green!20}Yes 
            & \cellcolor{yellow!20}-\\
        Articulating~\cite{der2021roboa} & \cellcolor{green!20}Nylon/TPU 
            & \cellcolor{red!20}No 
            & \cellcolor{red!20}No 
            & \cellcolor{green!20}Yes 
            & \cellcolor{green!20}Yes \\
        Localization~\cite{frias2023vine} & \cellcolor{red!20}LDPE 
            & \cellcolor{red!20}No 
            & \cellcolor{red!20}No 
            & \cellcolor{red!20}No 
            & \cellcolor{yellow!20}-\\
        Soft Cap~\cite{suulker2023soft} & \cellcolor{green!20}Nylon/TPU 
            & \cellcolor{green!20}Yes 
            & \cellcolor{green!20}Yes 
            & \cellcolor{red!20}No 
            & \cellcolor{yellow!20}- \\
        Large-Scale~\cite{heap2024large} & \cellcolor{red!20}Nylon/Silicone 
            & \cellcolor{red!20}No 
            & \cellcolor{green!20}Yes 
            & \cellcolor{green!20}Yes 
            & \cellcolor{yellow!20}- \\
        \textbf{Ours (Triangular)} & \cellcolor{green!20}Nylon/TPU 
            & \cellcolor{green!20}Yes 
            & \cellcolor{green!20}Yes 
            & \cellcolor{green!20}Yes 
            & \cellcolor{green!20}Yes\\
        \bottomrule
    \end{tabularx}
    \endgroup
\end{table}

\p{Body Material} Because vine robots grow through eversion, the body fabric must be able to pass freely through the tip mount. While all mounts introduce some resistive force, the magnitude of friction significantly depends on the fabric used in the vine body. Existing tip mount studies primarily employ three fabric types: silicone-coated nylon, low-density polyethylene (LDPE), and thermoplastic polyurethane (TPU)-coated nylon/kevlar. Silicone-coated fabrics tend to exhibit the lowest friction, making it easiest to design a successful tip mount for them. For example, we measured 1.1~oz MTN Silnylon 6.6 (Ripstop by the Roll) as having a coefficient of friction of 0.18. However, these robots are challenging to manufacture at scale, as the fabric has a bias and is sealed with glue. LDPE is a heat-sealable option that also offers low friction (friction coefficient 0.19~\cite{perez2022self}); however, it easily develops holes after repeated use. TPU-coated nylon, in contrast, can be heat-sealed and more easily manufactured at long lengths while remaining highly puncture-resistant in the field \cite{mcfarland2024field}, but it exhibits higher friction. We measured a friction coefficient of 0.84 for TPU-coated Ripstop-Nylon 6.6 (extremtextil) with its TPU-coated side contacting itself. Thus, it is a challenging, but worthy goal to develop a successful tip mount that works on TPU-coated fabric.

\p{Low Profile} When a vine robot encounters an opening smaller than its diameter, the body can constrict to pass through. The addition of a large rigid tip mount prevents this by imposing a larger diameter of constriction. Therefore, effective mounts should be soft or low-profile to minimize obstruction, such as one magnetically-attached cap \cite{luong2019eversion}, the soft cap \cite{suulker2023soft}, and the internal design \cite{heap2021soft}; although not soft, the internal mount fits within the robot body and still allows constriction. This property also affects how the mount interacts with the environment: because vine robots grow from the tip, the body remains stationary, and only externally attached mounts are dragged along the ground. Designs that do not attach to the exterior surface avoid this frictional interaction, which hinders motion.

\p{Lightweight} Vine robots are not able to carry excessive transverse loads, as they risk collapsing \cite{mcfarland2025modelingcollapsesteeredvine}. If a tip mount is too heavy, it may prevent the robot from lifting its tip and navigating effectively in free space. Most of these designs are heavy relative to the robot body's weight (>250 g). The magnetically-attached cap~\cite{luong2019eversion}, the the internal design~\cite{heap2021soft}, the soft cap~\cite{suulker2023soft}, and the large-scale variant on the internal design~\cite{heap2024large} all add minimal weight to the tip. 

\p{Secure} We consider the mount’s ability to resist forces that would pull it off the tip of the vine, dislodging it from alignment with the direction of growth. The securing strength of magnet-based designs~\cite{luong2019eversion, stroopa2020icra, stroppa2023shared} depends on the magnet size and the proximity of the mating surfaces. Hard~\cite{CoadRAM2020, frias2023vine} or soft~\cite{suulker2023soft} cap designs rely on friction with the robot body to stay on, which may be overcome by sufficiently directed external forces. More secure approaches include mechanical tethers \cite{mishima2006development}, which physically tie the mount to the body, and geometric constraint strategies \cite{heap2024large,jeong2020tip, heap2021soft}, which interlock two rigid pieces together.

\p{Fast} An ideal tip mount will allow a vine robot to grow at the maximum speed at which the body material may evert from a base. Our proposed design has a maximum growth speed of 3.65~m/min. The tool mount~\cite{jeong2020tip} had a speed of 3~m/min, and the hard cap~\cite{CoadRAM2020} and articulating design~\cite{der2021roboa} both achieved 6 m/min, albeit with nonidealities such as heavy weight or incompressible geometry. For most of the previous tip mount designs, speed was not a listed parameter, because it was not the focus of the work; these designs are marked in yellow in Table~\ref{tab:tip_mounts}. However, from video demonstrations, growth is often shown sped-up by a constant factor, illustrating the gap between the true performance and the desired speed. For urban search and rescue missions, where speed is critical, this is an important factor for any successful tip mount. 

\section{Tip Mount Design}
\label{sec:mount_design}
\begin{figure}[!b]
    \vspace{-1.5em}
    \centering
    \includegraphics[width = \columnwidth]{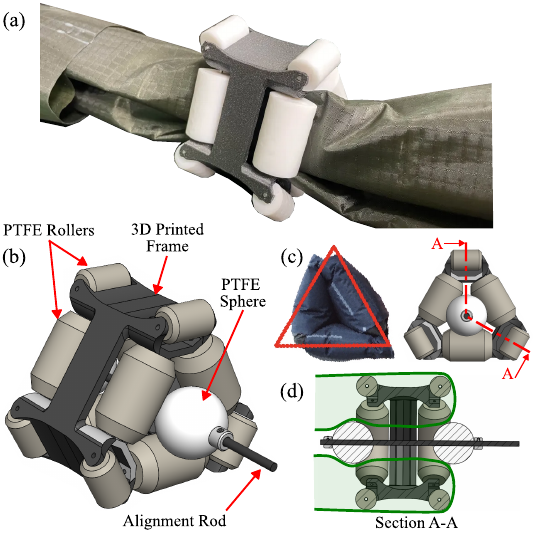}
    \caption{\textbf{\changed{High-speed} tip mount for vine robot.} (a) Mount installed on the vine body during the installation process. The fabric is wrapped up and around the body to finish the installation. (b) CAD Model highlighting key components: Teflon (PTFE) rollers, 3D printed frame, PTFE Spheres, and alignment rod. (c) Cross-sectional view of the triangular body profile (left) formed by the inflated vine and (right) the profile of the tip mount assembly. (d) Section A-A view showing internal and external rollers contact with the fabric and alignment of the PTFE spheres along the central rod.}
    \label{fig:mount}
\end{figure}
Our \changed{improved} tip mount design (Fig.~\ref{fig:mount}) is inspired by the internal tip mount design~\cite{heap2021soft}, which worked well on a silicone-coated nylon robot body. \changed{However, it} does not work on a TPU-coated nylon robot body. Our mount was developed to align mechanically with the natural deformation pattern of the vine robot body during pressurization and eversion. The robot body is constructed from 30-denier heat-sealable TPU-coated ripstop nylon (extremtextil) with three \textit{interior} series pouch-motor actuators integrated into the main chamber \cite{mack2025efficient, CoadRAM2020, mcfarland2025onsteerability}. When inflated, the body assumes a quasi-triangular cross-section due to the layout of the interior pouches and fabric seams (Fig.~\ref{fig:mount}(c)). This deformation inspired the development of a mount geometry that conforms to the triangular profile rather than opposing it. \changed{We posit} that a mount which complements the body’s inherent shape \changed{will distribute} internal pressure more evenly, \changed{minimize} local wrinkling, and \changed{ultimately} reduce the frictional contact area that contributes to drag during eversion.

The resulting mount consists of a 3D-printed structural frame housing two symmetric sets of rollers---one set facing inward toward the vine's internal surface and one facing outward toward the everting fabric. Each set contains three Teflon (PTFE) rollers equally spaced around the mount’s triangular perimeter. The inward-facing rollers are larger in diameter and length (14.00 mm x 31.25 mm) and positioned radially closer to the mount’s centerline. \changed{This placement} causes these rollers to remain in contact with the inner layer of the vine body (the fabric traveling toward the tip before eversion) and to roll against PTFE spheres at both ends. These internal rollers and spheres provide a self-centering bearing interface that maintains alignment of the mount rod with the direction of growth. The outward-facing rollers are smaller in diameter and length (9.53 mm x 19 mm) and slightly protrude from the mount’s exterior. These rollers contact the everting outer surface of the vine robot, guiding the fabric as it folds over the mount during growth. 

\begin{figure*}[!t]
    \centering
    \includegraphics[width=0.83\linewidth]{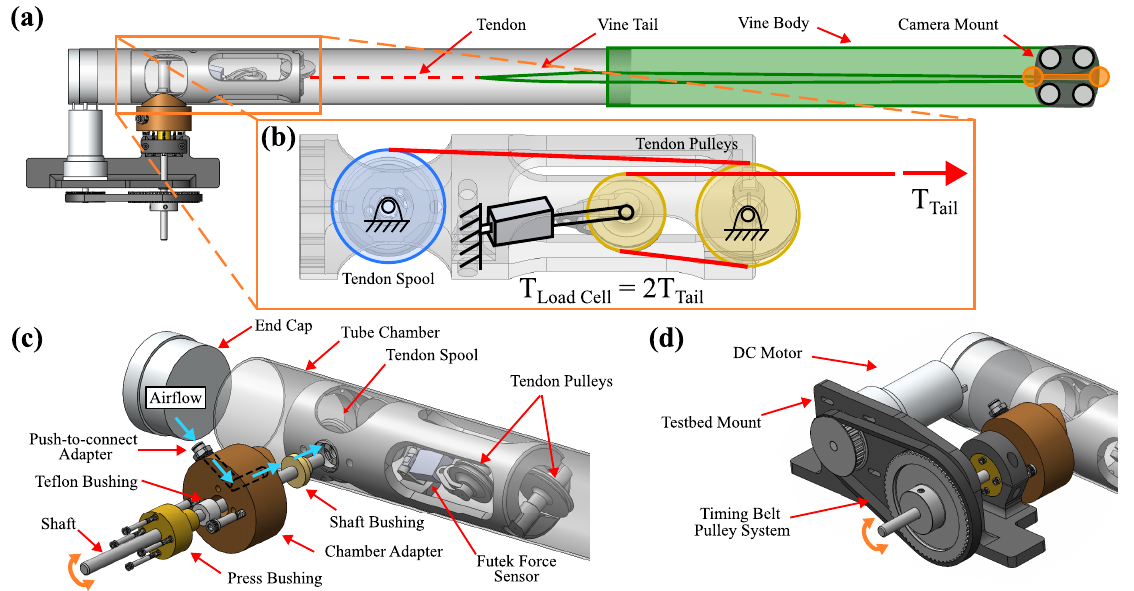}
    \caption{\textbf{Tip mount testbed design.} (a) Cross-section of vine growth and tail path in test bed. (b) Free-body diagram of forces on the load-cell and pulley under vine growth. (c) Isometric exploded view of the benchmarking testbed. (d) Drive system for testbed shaft.}
    \label{fig:testbed_figure}
    \vspace{-1.75em}
\end{figure*}
The frame is 3D-printed in polylactic acid (PLA) and fitted with machined PTFE rollers mounted on 1/16-inch stainless steel shafts. Each roller is fabricated from PTFE rod stock and drilled concentrically to allow free rotation around the shaft. The mount alignment rod consists of a 3~mm carbon fiber tube approximately 20~cm in length, fitted with shaft collars and PTFE spheres at both ends.
For assembly, one sphere is first secured to the rod, and the rod is passed through the 3D-printed frame while the larger internal rollers are installed on one side. The partially assembled mount is then inserted into the inverted vine body, and the second sphere and remaining large rollers are installed on the opposite side. Then, the smaller external rollers are mounted on both faces of the frame, matching the state shown in Fig.~\ref{fig:mount}(a). Finally, the vine body is folded over the mount to complete the installation, as shown in the diagram of the cross-section A-A in Fig.~\ref{fig:mount}(d).
The complete assembly weighs approximately 229.3~g and measures 69~mm x 69~mm x 82~mm.

\section{Testbed and Evaluation Metrics}
\label{sec:testbed}
Characterizing tip mounts on vine robots presents unique challenges that cannot be addressed using standard force testing systems. Conventional material testing systems can quantify static friction or tensile strength but cannot reproduce the coupled pneumatic and mechanical interactions that govern vine robot eversion. Prior studies have assessed tip mounts through task-based metrics---such as obstacle navigation performance \cite{suulker2023soft} or minimum pressure required for sustained growth \cite{jeong2020tip}---but these approaches provide only indirect measures of the mechanical interactions between the mount and the body. In particular, the dynamic process of eversion creates internal friction that varies with pressure, speed, and payload geometry between fabric layers and tip mounts, making it essential to measure the system in operation rather than under static loading. 

During growth, the internal pressure $P$ everts the vine body while tension at the tail $T_t$ resists this motion. A simplified force balance gives $T_t =$ \textonehalf $(PA-F_e)$, where $A$ is the tip cross-sectional area and $F_e$ represents the eversion force due to membrane bending and contact friction  \cite{coad2020retraction}. As $F_e$ increases, the measured $T_t$ decreases; thus, lower tail tension indicates higher frictional resistance during growth. To capture these interactions, we developed a custom pressurized testbed that integrates pneumatic control, tendon actuation, and tension sensing within a single apparatus. The system allows controlled inflation of the vine body while the spool motor drives growth and retraction, providing direct measurements of the net tensile load required to evert the structure and overcome frictional resistance introduced by the tip mount. 

For benchmarking, higher and more stable tail tension indicates improved growth efficiency and reduced friction from the mount, whereas irregular or spiking tension profiles correspond to jamming, internal drag, and geometric misalignment. \changed{We measure tension in a straight, horizontal setup to isolate the tip mount’s mechanical effect and ensure reproducibility. Prior work \cite{blumenschein2017modeling} shows tip friction primarily affects path-independent growth forces, justifying evaluation through a simplified testbed.} Thus, tail tension serves as a practical, quantitative performance \changed{basis} for a systematic evaluation to compare mount designs.

\subsection{Hardware Setup}
\label{sec:hardware_setup}

The testbed consists of a sealed cylindrical chamber designed to pressurize the vine robot body while providing direct mechanical actuation of the tail tendon. A schematic and annotated model of the system are shown in Fig.~\ref{fig:testbed_figure}.  The setup includes a custom 3D-printed internal assembly that houses a spool, a load-cell pulley system, and chamber interface components, all arranged concentrically within the tube chamber. The vine robot is attached to a tendon that is wound around a motorized spool, allowing growth and retraction to be controlled by motor commands. As illustrated in Fig.~\ref{fig:testbed_figure}(a), the tendon exits the chamber through the front opening so that it can be attached to the tail of the vine robot. 

The spool is mechanically driven by a DC motor (XD-60GA775 24V, Xinda Motor) located outside the airtight chamber (Fig.~\ref{fig:testbed_figure}(d)). This configuration is distinctive from most vine robot bases in the literature, where the actuation motor and coupling are housed within the pressurized chamber. \changed{This co-location} imposes geometric and volumetric constraints on the design. By contrast, our design locates all active mechanical components outside the chamber, allowing the chamber diameter to be reduced. The motor transmits torque to the spool through a timing-belt pulley system and operates at a fixed voltage, producing a consistent growth speed verified from timed extension. The shaft passes through the chamber wall via a machined aluminum adapter that serves as both a structural interface and an airtight feed-through.

The adapter performs three functions. First, it aligns the spool and shaft concentrically within the chamber using a machined bushing. Second, it integrates a channel machined directly through the aluminum body, connecting a push-to-connect fitting on the exterior to the chamber interior to deliver air from an external supply line. Third, it provides an anchoring surface for a press bushing that compresses a Teflon sealing bushing against the adapter. The Teflon bushing, chosen for its deformability and low-friction surface properties, conforms around the rotating shaft to create an airtight seal while allowing free rotation. The fasteners connecting the press bushing to the adapter are torqued tightly to ensure sufficient compression of the Teflon element, maintaining the chamber's air-tightness.

Inside the 3D-printed internal assembly, the spool is supported axially by a machined bushing that maintains alignment with the adapter and houses the tendon routing mechanism.  The tendon exits the spool and is routed through a two-pulley system, as shown in Fig.~\ref{fig:testbed_figure}(b). The right pulley is fixed in place and rotates about its shaft, while the left pulley is attached to a load cell and translates under tension while also rotating. The load cell is rigidly mounted to the internal frame, allowing direct measurement of the force transmitted through the tendon. Because the tendon wraps approximately $180^\circ$ around the pulley, the load cell measures twice the tail tension. The internal assembly is inserted into the chamber tube and fastened to the chamber adapter using machine screws. The entire chamber module is mounted on a rigid base plate (Fig.~\ref{fig:testbed_figure}(c)). This base can be clamped securely to a laboratory table using standard C-clamps. The CAD files and the bill of materials for the testbed are included in the supplementary material.

\begin{figure}[!tb]
    \centering
    \includegraphics[width = \columnwidth]{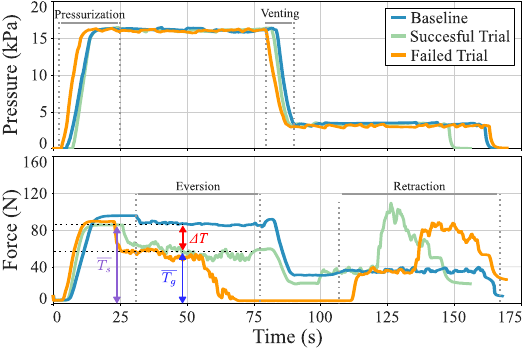}
    \caption{\textbf{Example testbed pressure and tail tension time series.} Pressure (top) and force (bottom) recorded during vine robot growth with and without a tip mount. The blue trace shows the \textit{baseline} condition (no mount), the green trace represents a \textit{successful trial} with a mount, and the orange trace corresponds to a \textit{failed trial} where the vine stalls mid-growth. There are four major operational phases---\textit{pressurization}, \textit{eversion}, \textit{venting}, and \textit{retraction}---to illustrate the temporal evolution of pressure and tail tension. During pressurization, tail tension rises to the static value $\overline{T_s}$ as the body becomes fully inflated. When eversion begins, it decreases to the steady growth value $\overline{T_g}$, primarily reflecting reduced internal and contact friction. The difference $\Delta T = \overline{T_s} - \overline{T_g}$ quantifies the additional frictional resistance introduced by the mount. An ideal mount reproduces the baseline profile, exhibiting minimal $\Delta T$ and stable growth.}
    \label{fig:force_plots}
    \vspace{-1.8em}
\end{figure}

\subsection{Experimental Protocol}
\label{sec:exp_protocol}

The testbed enables repeatable growth trials applicable to any vine–payload configuration. Each trial consists of a full pressurization–growth–retraction cycle inside the sealed chamber, performed under constant pressure and nominally constant spool speed. Prior to testing, all sensors and amplifiers are calibrated, and pneumatic, electrical, and mechanical connections are verified.

To prepare the robot, the spool tendon is first pulled out of the chamber tube and securely tied to the tail of the vine robot body. The spool motor is then activated to retract the robot into the chamber tube, leaving sufficient free length at the tip to install the desired tip mount. For the process followed to install the mounts in this work, please refer to Section~\ref{sec:mount_design}. After the mounts are installed, the vine robot is secured to the tube chamber using a hose clamp.

An air compressor (EC28M, Metabo HPT) supplies compressed air to a hand-operated pressure regulator (Type 700, ControlAir) connected to the chamber inlet. The regulator is set to provide 17.2~kPa during operation while the compressor maintains a supply pressure of 137.9~kPa. At the start of each run, the robot is fully retracted and depressurized to remove slack. The chamber pressure is then increased to the target value and held for several seconds to record steady-state tail tension under pressurization without growth. Subsequently, the spool motor is activated to initiate tip eversion at a nominal constant speed of $10$~mm/s. The robot grows to a predetermined length while time-synchronized pressure and tension data are continuously recorded. Upon reaching the desired extension, the motor is stopped, the chamber vented to approximately 4.8~kPa, and the robot retracted to complete the cycle. Three characteristic outcomes are observed, as illustrated in Fig.~\ref{fig:force_plots}. In the \textit{baseline} condition (blue traces), no mount is installed—the vine robot grows freely under pressurization, producing a smooth, repeatable tail tension profile that serves as the reference. When a tip mount is installed, trials can result in either \textit{successful} (green) or \textit{unsuccessful} (orange) outcomes based on growth continuity, as described in Section~\ref{sec:eval_metrics}.

All data are logged for post-processing and used to compute steady-state tail tension, maximum transient force, and trial-to-trial variability, which together quantify growth smoothness and reliability. The resulting pressure and force signals captured each major phase of operation, forming the basis for the analysis of tension dynamics described in Section~\ref{sec:eval_metrics}. This protocol is used in Section~\ref{sec:study} to compare different tip mount configurations under identical conditions.

\subsection{Data Acquisition and Processing}
\label{sec:data_acq}
During each trial, chamber pressure and tail tension data were collected using a microcontroller (Arduino Uno R4 Wi-Fi) at a serial rate of 115.2 kbaud. The internal chamber pressure was measured using an absolute pressure sensor (MPRLS, Honeywell). Tail tension was recorded using a load cell (FUTEK LSB200, 0–444.8 N range) connected to a signal conditioner (OMEGA DRF-LC) configured for a 0–10 mV input and a 0–10 V output. The conditioned signal was scaled to a 0–5 V range using a voltage divider and read through the Arduino’s 10-bit analog-to-digital converter (ADC). All signals were filtered using a first-order low-pass Butterworth filter to attenuate electrical noise.

\subsection{Evaluation Metrics}
\label{sec:eval_metrics}

As described in Section~\ref{sec:exp_protocol}, each trial produces synchronized pressure–force time series capturing the main operational phases: pressurization, eversion, venting, and retraction. The plots in Fig.~\ref{fig:force_plots} illustrate these phases for representative trials. The force traces correspond directly to the load-cell measurements, which record twice the actual tail tension because the tendon wraps approximately $180^{\circ}$ around the sensing pulley (see Section~\ref{sec:hardware_setup}). During pressurization, tail tension rises as internal pressure inflates the body while the spool remains stationary, reaching a static value $T_s$ that reflects the net pressure load on the vine body. When the spool motor initiates eversion, the tension decreases to a steady growth tension value $T_g$ as the inflated body begins to grow. This drop occurs because the effective friction and bending resistance are reduced once the body transitions from static inflation to continuous motion. The performance of each mount configuration is evaluated using the mean steady-state values $\overline{T_s}$ and $\overline{T_g}$ and their difference,

\begin{table*}[b]
\vspace{-1.0em}
\centering
\caption{\bf{Overview of Tested Tip Mount Configurations and Performance Outcomes}}
\label{tab:mount_results}
\renewcommand{\arraystretch}{1.25}

\begin{threeparttable}
    \resizebox{\textwidth}{!}{%
    {\large
    \begin{tabular}{clcccccc}
        \rowcolor[HTML]{EFEFEF}
        \textbf{ID} & \textbf{Mount Variant} & \textbf{Geometry / Design Feature} & \textbf{Mass (g)} &
        \textbf{Low Friction} & \textbf{Stable Growth} & \textbf{Repeatable} & \textbf{Outcome Summary}\\
        \hline
        (a) & Baseline (No Mount) & None — bare vine body & 0 &
        \cellcolor[HTML]{C7E9B4}\checkmark &
        \cellcolor[HTML]{C7E9B4}\checkmark &
        \cellcolor[HTML]{C7E9B4}\checkmark &
        \cellcolor[HTML]{C7E9B4}Reference for comparison\\
        
        (b) & Triangular (Inner + Front Spheres) & Triangular rollers with spheres inside and front & 120.3 &
        \cellcolor[HTML]{FEE6CE}\texttimes &
        \cellcolor[HTML]{C7E9B4}\checkmark &
        \cellcolor[HTML]{FEE6CE}\texttimes &
        \cellcolor[HTML]{FEE6CE}Consistent growth; moderate tension drop\\
        
        (c) & Triangular (Rear + Front Spheres) & Moved internal sphere to rear; kept front sphere & 120.3 &
        \cellcolor[HTML]{FEE6CE}\texttimes &
        \cellcolor[HTML]{C7E9B4}\checkmark &
        \cellcolor[HTML]{C7E9B4}\checkmark &
        \cellcolor[HTML]{FEE6CE}Smooth growth but high tension difference\\
        
        (d) & Triangular (Spring-Aligned) & Rear sphere with spring preload on front collar & 121.0 &
        \cellcolor[HTML]{FEE6CE}\texttimes &
        \cellcolor[HTML]{FEE6CE}\texttimes &
        \cellcolor[HTML]{FEE6CE}\texttimes &
        \cellcolor[HTML]{F4A6A6}Sticking during eversion; unstable\\
        
        (e) & Triangular (Front Roller Extension) & Added second roller assembly at front & 127.5 &
        \cellcolor[HTML]{F4A6A6}\texttimes &
        \cellcolor[HTML]{F4A6A6}\texttimes &
        \cellcolor[HTML]{F4A6A6}\texttimes &
        \cellcolor[HTML]{F4A6A6}Frequent jamming; excessive friction\\
        
        (f) & Circular Mount \cite{heap2021soft} & 3D-printed circular frame with bearings & 83.4 &
        \cellcolor[HTML]{F4A6A6}\texttimes &
        \cellcolor[HTML]{F4A6A6}\texttimes &
        \cellcolor[HTML]{FEE6CE}\texttimes &
        \cellcolor[HTML]{F4A6A6}Slow growth, inconsistent motion\\
        
        (g) & Short Triangular Mount & Compact single-row design & 77.9 &
        \cellcolor[HTML]{FEE6CE}\texttimes &
        \cellcolor[HTML]{F4A6A6}\texttimes &
        \cellcolor[HTML]{FEE6CE}\texttimes &
        \cellcolor[HTML]{F4A6A6}High friction due to tilting inside vine\\
        
        (h) & Hybrid Circular–Triangular & Circular rear section + triangular front rollers & 116.5 &
        \cellcolor[HTML]{FEE6CE}\texttimes &
        \cellcolor[HTML]{C7E9B4}\checkmark &
        \cellcolor[HTML]{C7E9B4}\texttimes &
        \cellcolor[HTML]{FEE6CE}Improved alignment; still high $\Delta T$\\
        
        (i) & PTFE Triangular & Triangular rollers machined from PTFE & 229.3 &
        \cellcolor[HTML]{C7E9B4}\checkmark &
        \cellcolor[HTML]{C7E9B4}\checkmark &
        \cellcolor[HTML]{C7E9B4}\checkmark &
        \cellcolor[HTML]{C7E9B4}\textbf{Smoothest growth; lowest $\Delta T$}\\
        \hline
    \end{tabular}%
    } 
    } 
    \begin{tablenotes}
    \small
    \item \textit{Color Key:} 
    \cellcolor[HTML]{C7E9B4}\textbf{Green} = successful/consistent;
    \cellcolor[HTML]{FEE6CE}\textbf{Orange} = partial/inconsistent;
    \cellcolor[HTML]{F4A6A6}\textbf{Red} = failure/jamming.
    \end{tablenotes}
\end{threeparttable}
\end{table*}

\begin{equation}
    \Delta T = \overline{T_s} - \overline{T_g},
\end{equation}
which quantifies the reduction in tail tension between static pressurization and active growth. \changed{We note that $\Delta T$ represents an aggregate measure of eversion resistance, capturing the combined effects of membrane bending, internal sliding between fabric layers, and contact friction between the tip mount and the robot body, rather than a direct measurement of friction alone.} 
From the force balance in Section~\ref{sec:testbed}, $\Delta T$ corresponds to half of the effective eversion force $F_e$, which includes both membrane bending and contact friction. Thus, the increase in $\Delta T$ value with a tip mount, as compared to the value without one, reflects how the mount alters the net mechanical resistance opposing growth. Smaller $\Delta T$ values indicate smoother eversion and lower friction resistance, while large or fluctuating $\Delta T$ values correspond to increased drag or mechanical instability. The steady-state values $\overline{T_s}$ and $\overline{T_g}$ are obtained by identifying regions of low temporal variance in the filtered tension signal. Specifically, $\overline{T_s}$ was computed over the stable plateau following pressurization, and $\overline{T_g}$ was calculated over the subsequent interval between the detected pressure drop and the end of the eversion phase. Trials that exhibited uninterrupted growth were classified as \textit{successful}, while those showing a temporary loss of motion---often caused by mount sticking or internal friction---were labeled as \textit{failures}. These events appear as sharp tension drops ($T_g \rightarrow 0$) in the force signal, after which the motor was briefly paused to allow the tip to re-engage. Failed trials were excluded from steady-state averaging, although valid growth segments preceding the stall were retained for mean-tension analysis. 

\section{Comparative Study of Mounts}
\label{sec:study}
\begin{figure*}[!t]
\centering
\includegraphics[width=0.95\linewidth]{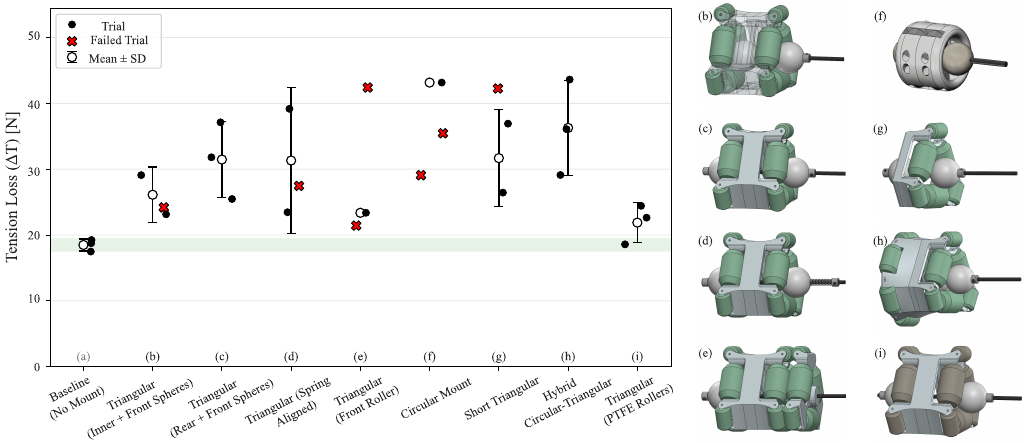}
\caption{\textbf{Effect of tip mount design on tension loss ($\Delta T$, defined in Fig.~\ref{fig:force_plots}) during vine robot growth.} Each point represents an individual trial; red markers denote failed growth attempts, and open circles with bars indicate mean~$\pm$~standard deviation for successful trials. The green shaded region shows the baseline range (no mount). Tip mount configurations (b–i) correspond to the mount geometries shown on the right, illustrating the design iterations from early triangular assemblies to the \changed{final} PTFE-roller mount, configuration (i).}
\label{fig:results}
\vspace{-1.8em}
\end{figure*}

To evaluate how internal mount design choices affect eversion on our TPU-coated, quasi-triangular vine body, we conducted a comparative study within the internal-mount family. We systematically varied alignment hardware (sphere placement, spring preload), contact layout (roller rows/length), and surface friction (printed polymer vs. PTFE rollers), while holding pressure, growth speed, and body material constant. For reference, we also included a representative circular internal mount adapted from prior literature~\cite{heap2021soft}. All variants were benchmarked on the same testbed using the metrics in Section~\ref{sec:eval_metrics} (tension loss $\Delta T$, stability across trials, and deviation from the baseline no-mount condition).

\subsection{Mount Variants}
\label{sec:mount_variants}
We developed eight internal-mount variants (Fig.~\ref{fig:results}(b–i) and Table~\ref{tab:mount_results}) to probe the internal contact mechanics most relevant to our body: alignment strategy, preload, roller placement/extent, and material friction, while keeping mass and stiffness comparable. All mounts used 3D-printed PLA frames with stainless steel shafts, and incorporated low-friction PTFE components where applicable. 

Mounts \textit{b–e, g, i} adopted a \textit{triangular geometry} consistent with the vine’s quasi-triangular inflated profile (Section~\ref{sec:mount_design}). The \textit{Inner + Front Spheres} configuration introduced a pair of PTFE spheres to maintain self-centering alignment during eversion. The \textit{Rear + Front Spheres} variant repositioned the internal sphere to the back of the frame to reduce internal fabric friction and promote smoother flow. The \textit{Spring-Aligned} design added a compression spring between the front sphere and collar to apply a constant preload and accommodate fabric thickness changes during growth. The \textit{Front Roller Extension} incorporated an additional roller array at the mount tip to increase contact length and distribute pressure more uniformly along the fabric. The \textit{Triangular (PTFE Rollers)} configuration, our \changed{improved} design from Section~\ref{sec:mount_design}, replaced the 3D-printed rollers with machined PTFE rollers to minimize friction at both fabric interfaces, forming the final design. A shortened \textit{Triangular (Short)} variant reduced structural length and mass while testing the limits of self-centering stability. Two additional designs examined cross-sectional effects. The \textit{Circular Mount}, adapted from prior work~\cite{heap2021soft} is a 3D-printed circular frame with bearings. The \textit{Hybrid} version replaced the rear triangular rollers with a six-roller circular array to balance geometric conformity with smoother radial contact. The following subsections present quantitative comparisons of tension loss $\Delta T$ and configuration-specific observations.

\subsection{Quantitative Results}
\label{sec:results}
Fig. \ref{fig:results} summarizes the measured tension loss $\Delta T$, for all tip mount configurations tested. The baseline condition (no mount) exhibited the lowest and most consistent $\Delta T$ ($18.5 \pm 0.9$ N), defining the green performance band used for comparison. All mounted configurations produced larger $\Delta T$ than the baseline, reflecting the additional friction introduced by the mount assemblies.

Among the prototypes, the \changed{final} mount (PTFE-rollers in a triangular shape) achieved the smallest deviation from the baseline ($21.9 \pm 3.0$ N) and exhibited smooth, repeatable growth across all trials. Designs featuring internal spheres or extended roller assemblies showed higher mean $\Delta T$ and greater variability, often accompanied by partial or full jamming events, annotated as a red mark in Fig.~\ref{fig:results}. 
These results indicate that both geometric conformity to the inflated body and low-friction contact surfaces are critical to minimizing eversion resistance. The following subsections qualitatively analyze the observed mechanical behavior of each configuration and link these trends to the quantitative outcomes presented in Fig. \ref{fig:results} and Table \ref{tab:mount_results}.

\subsection{Configuration-Specific Observations}
\p{Triangular Designs} The triangular mount configurations generally produced smooth, stable growth and moderate $\Delta T$ values, with performance variations tied to alignment and friction characteristics. The \textit{Inner + Front Spheres} mount yielded consistent results ($\Delta T = 26.1 \pm 4.2$ N) but exhibited slightly elevated friction due to internal sphere contact. Relocating this sphere to the rear (\textit{Rear + Front Spheres}) improved alignment but increased overall drag ($\Delta T$ = $31.5 \pm 5.8$ N), suggesting added resistance at the fabric–sphere interface. The \textit{Spring-Aligned} concept maintained proper preload but intermittently jammed, producing the highest variability ($31.3 \pm 11.1$ N) among the triangular designs. The \textit{Front Roller Extension} design showed high contact drag, jamming, and incomplete growth, with only one successful trial ($23.4$ N) and two failures. Replacing 3D-printed rollers with machined \textit{PTFE rollers} for the \changed{final} design substantially reduced friction and achieved smooth, repeatable growth across trials ($21.9 \pm 3.0$ N), approaching baseline performance. The \textit{Short Triangular} mount reduced weight but was prone to tilting inside the inflated body, causing asymmetric contact and higher variability ($31.7 \pm 7.4$ N, one failure). These results confirm that sufficient structural length and low-friction materials are critical for maintaining stable self-centering growth.

\p{Circular and Hybrid Designs} The circular-based configurations exhibited higher $\Delta T$ values and more frequent failures. The \textit{Circular Mount}, adapted from prior work \cite{heap2021soft}, recorded the largest tension losses ($43.2$ N) and multiple stalls, attributed to geometric incompatibility with the vine's triangular profile. The \textit{Hybrid Circular–Triangular} design reduced stalling frequency but still showed substantial resistance ($36.3 \pm 7.3$ N), reflecting residual misalignment and increased surface drag from its extended contact area.

\section{Discussion}

\p{Influence of Geometry on Growth Dynamics}
The data in Fig.~\ref{fig:results} confirm that the mount’s cross-sectional geometry plays a dominant role in frictional behavior. Circular mounts produced the largest tension losses and exhibited frequent growth failures, as reflected by the number of failed trials. These results support the hypothesis that circular geometries are incompatible with this vine’s triangular inflation pattern, which causes uneven pressure distribution and localized fabric constriction. In contrast, the triangular mounts that conformed more closely to the inflated body shape provided smoother, more repeatable growth and tension losses approaching the baseline condition. Because the masses of all mounts were comparable (Table~\ref{tab:mount_results}), these differences are attributed primarily to geometric and contact effects rather than weight. Another contributing factor is the available rolling surface: the circular mount from the literature uses two narrow bearing rows, providing limited circumferential contact, whereas the triangular designs incorporate larger and wider rollers that substantially increased the total rolling interface area. 

\p{Material and Surface Friction Effects} Material selection also proved critical. Substituting 3D-printed polymer rollers with machined PTFE substantially reduced $\Delta T$ and eliminated failures across repeated trials. While some of this improvement may be attributed to differences in geometry, the low friction of PTFE clearly facilitated fabric motion. In practice, the rollers did not behave as ideal bearings---local variations in fabric tension and contact pressure often caused intermittent sticking or partial sliding. As a result, the PTFE rollers likely alternated between true rolling and low-friction sliding, collectively reducing resistance compared with the higher-friction printed polymer surfaces.

\p{Alignment Stability and Load Distribution} Configurations incorporating external spheres, springs, or shortened frames reveal the importance of maintaining axial alignment. Although these features were introduced to improve self-centering, their performance depended strongly on the mechanical balance between constraint and freedom. The \textit{Spring-Aligned} and \textit{Short Triangular} designs, for instance, exhibited intermittent jams and large tension variability, suggesting that insufficient structural length or excessive compliance can destabilize the internal load path.

\p{Design Guidelines} From the collective outcomes, several design principles emerge:
\begin{enumerate}
    \item Match the cross-section geometry to the inflated body shape to minimize internal resistance.
    \item Use low-friction materials (e.g., PTFE) at all contact surfaces and increase the total rolling interface area to prevent fabric from sticking.
    \item Avoid over-constraining alignment; excessive structural elements (e.g., dual spheres or spring preloads) can introduce new sources of friction.
    \item Maintain sufficient mount length for passive self-centering while preventing tilt-induced asymmetries.
\end{enumerate}

\p{Toward Standardized Benchmarking}  
Beyond design \changed{improvements}, the presented testbed and benchmarking protocol address a broader issue in the field: the lack of reproducible methods for quantifying tip mount performance. Previous studies often reported qualitative growth success or single-trial results, which conceal the impact of mechanical parameters such as pressure, speed, and payload geometry. In this work, we isolate tail tension as a measurable indicator of eversion efficiency, establishing a repeatable framework that enables direct comparison across laboratories and mount designs. The release of complete testbed and mount specifications online supports this goal of community-level standardization.

\p{Limitations and Future Work} While the $\Delta T$ metric captures the aggregate effect of mount–fabric interactions, it does not fully distinguish between internal sliding friction, bending losses, and external drag at the mount surface. Future work should combine tension sensing with local pressure and motion measurements to resolve these contributions. Preliminary tests also showed that, for the baseline (no tip mount), tail tension decreases with increasing growth speed---an effect likely linked to dynamic frictional behavior. This observation suggests that the testbed can be used to systematically investigate the influence of eversion speed on frictional performance across materials and mount geometries. \changed{The evaluation in this work focuses on axial growth in a controlled setting to isolate the mechanical effects of tip mount design. While this enables repeatable benchmarking, it does not fully capture behaviors in confined or cluttered environments; qualitative demonstrations of these scenarios are provided in the supplementary video.}

The testbed design additionally enables the development of lighter, more compact vine robots by allowing electromechanical assemblies to be moved outside of the pressurized portion. Mechanical improvements to the mount itself remain an important area of investigation. Currently, the central shaft can rotate relative to the outer rollers. To mitigate this, we plan to machine a specialized geometry to replace the Teflon balls, ensuring proper alignment and preventing unintended rotation, which is a feature critical for ego-centric visual sensing and teleoperation in confined spaces \cite{whitman2018snake}. Additional avenues for future work include designing mounts that resist engulfment during retraction and implementing reliable wiring solutions for delivering power and transmitting signals to the camera. 

\section{Conclusion}

In this paper, we presented an integrated framework for transporting tip-mounted payloads on growing vine robots. Our triangular roller tip mount reduces internal friction by matching the robot’s natural triangular deformation during growth, enabling faster and more reliable operation. Using a custom-built benchmarking testbed, we quantitatively demonstrated that this design achieves the lowest tail tension and most consistent eversion performance among the tested variants and prior state of the art in internal tip mounts.

Together, these contributions establish both a validated mount architecture and a standardized method for comparing future designs. \changed{While our study focused on complete mount assemblies, a systematic design-of-experiments approach could further explore the effects of individual design parameters and interactions, guiding continued optimization.} Building on this foundation, we plan to deploy the tip mount design in urban search and rescue field studies to carry sensing payloads and map the void spaces within rubble piles. This combination of mechanical innovation and quantitative benchmarking advances the field toward vine robots capable of rapid, sensor-integrated deployment in confined or hazardous environments, providing critical visual and spatial information for first responders.

\bibliographystyle{IEEEtran}
\bibliography{library}

\end{document}